# Predicting Mortality and Functional Status Scores of Traumatic Brain Injury Patients using Supervised Machine Learning

Lucas Steinmetz
Dept. of Computing Sciences
Villanova University
lsteinme@villanova.edu

Shivam Maheshwari
Unionville High School
maheshwaris1231@gmail.com

Garik Kazanjian
Dept. of Biology
Villanova University
gkazanji@villanova.edu

Abigail Loyson
Dept. of Computing Sciences
Villanova University
aloyson@villanova.edu

Tyler Alexander
Thomas Jefferson University
tyler.alexander@students.jefferson.edu

Venkat Margapuri
Dept. of Computing Sciences
Villanova University
vmargapu@villanova.edu

C. Nataraj
Dept. of Mechanical Engineering
Villanova University
c.nataraj@villanova.edu

*Abstract*—
Traumatic brain injury (TBI) is a major public health concern that often leads to mortality or long-term disability. Accurate prediction of outcomes, such as mortality and Functional Status Scale (FSS) scores, improves treatment strategies and guides clinical decisions. This study applies supervised machine learning (ML) techniques to predict mortality and FSS score in a real-world dataset of 300 pediatric TBI patients compiled by the University of Colorado School of Medicine. The dataset includes detailed clinical information, such as demographics, injury mechanisms, and hospitalization outcomes. Eighteen ML models predict mortality, while thirteen models forecast FSS scores. Performance metrics, including accuracy, ROC AUC, F1-score, and mean squared error, evaluate the models. Logistic regression and Extra Trees show high precision in mortality prediction, while linear regression achieves the best FSS score prediction with an R² score of 0.62. Feature selection reduces 103 clinical variables to the most relevant ones, improving model efficiency and interpretability. The study highlights how ML models help clinicians identify high-risk cases and support personalized interventions. This research demonstrates the potential of ML-driven predictive analytics to integrate into clinical workflows, providing healthcare professionals with data-driven tools to improve TBI care.

## I. INTRODUCTION

Traumatic Brain Injury (TBI) is a common diagnosis with over 200,000 related hospitalizations per year in the United States, of which, roughly 30% result in mortality [1]. For a variety of reasons, the incidence of TBI is on the rise both within the US and globally [2]. The severity of a TBI can vary depending on the mechanism of injury and includes concussions as well as high-impact trauma. Patients that suffer from TBI may be placed under intensive care depending on the severity of their injury. Management of a patient with a TBI requires the healthcare provider to integrate a combination of electronic health record (EHR) and clinical data in decision making. In certain cases of TBI, providers will want to be able to appropriately judge the risk of patient mortality – and in many other cases, judge the risk of long-term complications. A disabling consequence of TBI seen in more than 50% of the patients is post-concussion fatigue [3]. The Fatigue Severity Scale (FSS) is used to assess the impact on daily functioning the fatigue has on the patient. Physicians use a questionnaire containing nine questions evaluated on a seven-point scaling system wherein higher scores correlate with greater fatigue. Depending on the severity of the TBI, being able to predict patient outcomes plays an important role in the management of those patients. However, predicting patient outcomes upon being afflicted by a TBI remains challenging. It is due to the presence of thousands of data points which can impact mortality or the decision-making process behind selecting the appropriate treatment [4]. Although the human brain is well adjusted to data analysis, it struggles to comprehend datasets comprising a combination of linear and non-linear data points. As a result, predicting TBI outcomes is challenging even for the best physicians.

Machine learning driven systems have the potential to significantly improve the decision-making process of the physicians by providing insight into features that the human brain does not comprehend easily. They provide us with a diverse array of computational methods that can harness both linear and nonlinear features of the dataset [5][6][7][8][9][10][11]. In general, the management strategies for TBI can be greatly impacted if patient outcomes can be accurately predicted. Machine learning provides a potent arsenal of tools for such predictions, each with unique advantages. In particular, the significance of advancing predictive analytics for TBI patients is that it can improve treatment strategies. Complex medical datasets can be overwhelming in clinical settings; therefore, machine learning models aid healthcare professionals in making well informed decisions about TBI considering patient mortality and FSS score, enhancing the overall quality of care for patients. Despite recent advancements in machine learning, there is a notable gap in applying machine learning models within clinical settings for TBI. Few studies have explored comprehensive approaches that combine datasets comprising linear and non-linear features to accurately predict mortality and functionality status [12][13][14]. A holistic approach to solving this problem can improve patient care standards and practices in hospital settings. In addition, such models can be integrated in emergency rooms where physicians are able to leverage predictive analytics to assign treatments to patients diagnosed with a TBI.

The authors hypothesize that machine learning techniques provide an accurate and comprehensive analysis of traumatic brain injury patients by predicting mortality and functionality scores. Traditional statistical methods are not adept at finding subtle patterns, especially in data that are heterogeneous such as a combination of clinical and physiological data. Machine learning models can integrate multiple types of data to predict outcomes with a high degree of precision. Furthermore, machine learning algorithms can be personalized and tailored to individual patient needs as they uncover their injury profiles. The key contributions of the paper are:

1. The application of supervised machine learning techniques for the prediction of mortality and FSS score on a real-world TBI dataset comprising 300 pediatric patients.
2. The extraction and curation of key features for the tasks of mortality and FSS score prediction for TBI.
3. Elaborate experiment and comparative analysis using multiple performance metrics to demonstrate the feasibility and clinical relevance of supervised machine learning algorithms for TBI related use cases of mortality and FSS score prediction.

## II. RELATED WORK

Supervised machine learning has emerged as a transformative approach in the field of TBI by enhancing diagnostic and predictive capabilities. Supervised machine learning algorithms are increasingly utilized to predict patient outcomes, including mortality and functional status scores, based on diverse clinical data and imaging results. Furthermore, machine learning models aid in risk stratification [54], enabling healthcare providers to identify patients requiring urgent interventions. These algorithms also facilitate automated diagnosis by analyzing imaging data, such as CT scans, thereby improving diagnostic accuracy and informing personalized treatment strategies [55]. In the context of TBI, both classification and regression algorithms play crucial roles in predicting outcomes like mortality and functional status scores. The key works that inspired the selection of the machine learning algorithms for the current work are discussed below.

Steyerberg et al. [54] employed Logistic Regression [22], a classification approach, to accurately predict in-hospital mortality based on clinical parameters, highlighting the importance of age and injury severity. In contrast, Moore et al. [58] utilized linear regression to analyze the incidence of acute kidney injury in patients with TBI. These studies underscore the complementary nature of classification and regression in enhancing TBI patient management.

Bellotti et al. [55] presented a methodology aimed at investigating magnetic resonance imaging (MRI) scans of subjects after a TBI to detect the presence of heterogenous lesions and access the information content within using graph and K Nearest Neighbors (KNN) algorithms. The results showed that topological measurements provide a suitable measurement to detect and distinguish features between different subjects. In the same regard, Ball Tree [17] and KD Tree [30] algorithms have shown promise in medical applications efficiently handling high-dimensional data, such as neuroimaging and clinical metrics. They enable rapid nearest neighbor searches facilitating the classification and clustering of patient data to identify patterns and predict outcomes.

Petrov et al. [56] applied the gradient boosting [16][29] and Random Forest [31] machine learning algorithms in a study involving 36 patients admitted to a level I trauma center with severe TBI wherein the objective was to minimize secondary brain injury related to elevated intracranial pressure (ICP). Accurate prediction of ICP crises allows clinicians to implement preventative strategies to avoid further brain damage. The results showed that the random forest algorithm yielded the best outcomes in terms of accuracy, precision, recall, and F1 score.

Wang et al. [57] performed the prediction of acute respiratory distress syndrome in TBI patients using the machine learning algorithms of extreme gradient boosting [29], Random Forest [31], Adaptive Boosting [16], Naive Bayes [24], and Support Vector Machine [28] on data belonging to 649 TBI patients from the Medical Information Mart for Intensive Care – III (MIMIC-III) database. In the experiment where area under the curve was used as the evaluation criteria, the Random Forest algorithm provided the best result followed by the AdaBoost and XGBoost algorithms.

Thatcher et al. [58] conducted electroencephalogram spectral analyses from 19 scalp locations for patients with mild, moderate, and severe TBI, 15 days to four years after injury. The severity of TBI was assessed by GCS, and duration of coma and amnesia. The machine learning technique of multivariate discriminant analysis was applied to understand the correlations among the different groups of patients. In an experiment on 503 patients, significant correlations between EEG discriminant scores, emergency admissions procedures, and post-trauma neuropsychological test scores validated the discriminant function developed using the multivariate discriminant analysis technique as an index of severity of injury and classifier of the extremes of severity. The metrics of accuracy, precision, and recall used for the experiment showed the results of 96.3%, 95.4%, and 97.4% respectively.

## III. DATASET AND METHOD

### 3.1. Data Population

The study is conducted using TBI data for pediatric patients collected by the University of Colorado School of Medicine. The study population consists of 300 pediatric patients, with a dataset that includes comprehensive information such as age, gender, mechanism of injury, and various clinical interventions and outcomes. It also assesses the severity of TBI using measures such as the Glasgow Coma Scale (GCS) and includes relevant laboratory results. Each patient's hospitalization outcome—whether they survive or not—is documented. The age range of the cohort is from 5 days to 17.8 years, with an average age of 7.23 years and a standard deviation of 5.46 years. The inclusion criteria focus on the outcome of hospitalization, requiring patients to be either in intensive care or in the operating room. Furthermore, cardiovascular medical records prior to and during hospitalization are considered. Approximately 16% or 47 of the 300-patient clinical stays in the hospital resulted in mortality, with the remaining patients being discharged from the hospital. The mean FSS score across all domains was 1.7, with a maximum score of 2.0 for motor function and a minimum score of 1.2 for response. The dataset is publicly available for review and research purposes at https://github.com/marven22/TraumaticBrainInjuryDetection.

3.2 Research Questions

The experiment described in the paper to predict mortality and FSS scores in TBI patients using supervised machine learning aims to answer the following research questions (RQ):

- RQ-1: How do supervised machine learning algorithms compare amongst themselves to predict mortality in TBI patients?
- RQ-2: How do supervised machine learning algorithms compare amongst themselves to predict FSS scores in TBI patients?
- RQ-3: How do the features in the dataset relate to the prediction of mortality and FSS scores?
- RQ-4: What is the clinical relevance of the results in guiding the selection of one algorithm over another?

*3.3 Machine Learning Algorithms*

A total of eighteen supervised machine learning algorithms are employed to predict mortality rates, while thirteen are utilized to forecast Functional Status Scale (FSS) scores for traumatic brain injury (TBI). Table I presents a comprehensive list of the models used for predicting mortality rates and FSS scores. The wide array of models helped capture diverse aspects of the data as each model employs different computational approaches.

**Table I:** Supervised Machine Learning Algorithms for Mortality and FSS Prediction

| Ada Boost [16] | Logistic Regression [22] | Support Vector Machine (SVM) [28] |
|---|---|---|
| Ball Tree [17] | Multi-layer Perception (MLP) [23] | XGBoost [29] |
| Gradient Boosting [18] | Naïve Bayes [24] | KDTree [30] |
| Hist Gradient Boosting [19] | Nearest Centroid [25] | Random Forest [31] |
| K-Nearest Neighbors (KNN) [20] | Nearest Neighbors [26] | Radius Neighbors [32] |
| Extra Trees [21] | Linear Discriminant Analysis [27] | Quadratic Discriminant Analysis [33] |

*3.4 Performance Evaluation Metrics*

Predicting mortality rate is categorized as a classification problem, whereas forecasting FSS scores is treated as a regression problem. Consequently, different evaluation metrics are employed to assess the performance of machine learning models, tailored to each use case. Table II outlines the various performance metrics utilized for evaluating these models.

**Table II:** Performance Metrics

| Metric | Description | Mortality or FSS Score |
|---|---|---|
| Receiving Operating Characteristic Area Under the Curve (ROC AUC) [36] | Evaluates the model's ability to differentiate between positive and negative cases across thresholds. | Mortality |
| Accuracy [37] | Measures the proportion of correctly predicted instances out of the total instances. | Mortality |
| F1 Score [38] | Balance between precision and recall scores | Mortality |
| Mean Squared Error [35] | Quantifies the average squared difference between the actual and predicted values (lower values indicate better performance) | FSS Score |
| R-Squared ($R^2$) Error [39] | Measures the proportion of the variance in the dependent variable that is predicted from independent variable | FSS Score |
| Root Mean Squared Error [40] | Square root of means squared error | FSS Score |
| Mean Absolute Error [42] | Average absolute difference between the actual and predicted values | FSS Score |

*3.5 Experiment*

The performance of each of the machine learning models on the tasks of mortality prediction and FSS score prediction in TBI patients are shown in Table III and Table IV respectively. To make the most of the limited data available for the experiment and ensure that the machine learning models can effectively generalize to unseen data, the K-Fold Cross Validation technique [34] is employed. This approach enhances the reliability and consistency of the evaluation for each model, addressing concerns about their generalizability. The dataset is divided into five folds, with models trained on four independent folds and tested on the remaining fold during each iteration. The metrics documented in the subsequent sections represent the average results obtained across all iterations, providing a comprehensive assessment of model performance.

**Table III:** Model Performance on Mortality Prediction

| Model | Accuracy (%) | Precision | Recall | F1 Score | ROC AUC |
|---|---|---|---|---|---|
| AdaBoost | 94.66 | 0.87 | 0.79 | 0.81 | 0.96 |
| Ball Tree | 83.33 | 0.45 | 0.06 | 0.10 | 0.61 |
| Extra Trees | 96.66 | 0.98 | 0.81 | 0.88 | 0.96 |
| Gradient Boosting | 96.33 | 0.91 | 0.85 | 0.88 | 0.97 |
| Hist Gradient Boosting | 96.33 | 0.94 | 0.81 | 0.87 | 0.96 |
| KDTree | 96 | 0.95 | 0.78 | 0.85 | 0.94 |
| KNN | 95.33 | 0.94 | 0.77 | 0.83 | 0.93 |
| Linear Discriminant Analysis | 96 | 0.95 | 0.79 | 0.85 | 0.96 |
| Logistic Regression | 97 | 0.98 | 0.83 | 0.90 | 0.96 |
| MLP | 94 | 0.83 | 0.83 | 0.81 | 0.93 |
| Naive Bayes | 88.66 | 0.68 | 0.53 | 0.59 | 0.94 |
| Nearest Centroid | 94 | 0.79 | 0.85 | 0.81 | 0.81 |
| Nearest Neighbors | 95.33 | 0.96 | 0.75 | 0.82 | 0.91 |
| Quadratic Discriminant Analysis | 88.33 | 0.61 | 0.89 | 0.71 | 0.96 |
| Radius Neighbors | 84.33 | 0.8 | 0.04 | 0.08 | 0.51 |
| Random Forest | 96 | 0.95 | 0.79 | 0.85 | 0.96 |
| SVM | 95.66 | 0.87 | 0.86 | 0.85 | 0.97 |
| XGBoost | 96.33 | 0.91 | 0.85 | 0.88 | 0.97 |

**Table IV:** Model Performance on FSS Score Prediction

| Model | $R^2$ Score | Root Mean Squared Error | Mean Squared Error | Mean Absolute Error |
|---|---|---|---|---|
| AdaBoost | 0.35 | 3.31 | 11.43 | 2.62 |
| Ball Tree | -0.20 | 4.56 | 21.53 | 3.46 |
| Extra Trees | 0.03 | 6.17 | 39.05 | 3.43 |
| KDTree | -0.2 | 4.56 | 21.53 | 3.47 |
| KNN | -0.09 | 7.9 | 63 | 5.50 |
| Linear Regression | 0.62 | 3.62 | 13.1 | 2.52 |
| Logistic Regression | 0.24 | 3.63 | 13.87 | 2.23 |
| MLP | -0.007 | 7.56 | 57.57 | 5.17 |
| Nearest Neighbors | -0.18 | 8.17 | 66.97 | 5.88 |
| Radius Neighbors | 0.026 | 6.39 | 41.28 | 4.62 |
| Random Forest | 0.05 | 7.31 | 54.06 | 4.20 |
| SVM | -0.05 | 7.67 | 60.67 | 4.04 |
| XGBoost | -0.03 | 7.57 | 58.47 | 4.32 |

*3.6 Evaluation*

The results from the experiment are used to answer the three RQ mentioned in section 3.2.
1. How do supervised machine learning algorithms compare amongst themselves to predict mortality in TBI patients?

In medical applications, selecting the best model (algorithm) requires balancing various metrics to meet clinical needs. While accuracy is a general measure of correctness, it can be misleading for imbalanced datasets. In the dataset considered, mortalities represent only a minor percentage (16%) of cases. Precision is crucial when false positives are costly, such as recommending unnecessary procedures, and recall is critical when missing positive diagnoses is detrimental to the application. A comprehensive observation shows that Logistic Regression and Extra Trees achieve the best precision of 0.98, demonstrating that they seldom yield false positives. Quadratic Discriminant Analysis achieves the highest recall with 0.89 making it ideal when identifying all the positive cases is a priority, even though its precision is lower (0.61), meaning it may result in numerous false positives. When both precision and recall are critical, the F1 Score is considered the key metric as it balances these competing demands. Logistic Regression achieves the highest F1 Score (0.90), rendering it suitable when both misdiagnosis and missed diagnoses carry significant risks. XGBoost and Gradient Boosting also offer competitive F1 Scores (0.88) with a good balance of recall (0.85) and precision (0.91). Finally, the metric of ROC AUC is critical for models where flexibility in adjusting thresholds is essential providing insight into their overall discriminative abilities. SVM, XGBoost, and Gradient Boosting achieve the highest ROC AUC (0.97) indicating they perform well across different thresholds and can be fine-tuned based on clinical needs.

2. How do supervised machine learning algorithms compare amongst themselves to predict FSS scores in TBI patients?

The analysis of regression algorithms to predict the FSS scores shows significant variability in performance as indicated by $R^2$ Score, RMSE, MSE, and MAE. The $R^2$ Score measures the proportion of variance explained by the algorithm, with higher values indicating better predictive capability. The results show that Linear Regression stands out with the highest $R^2$ Score of 0.62, suggesting it effectively captures the variability in FSS scores. Furthermore, its RMSE of 3.62 indicates the average distance between predicted and actual values, while the MAE of 2.52 reflects the average absolute error, both suggesting that the algorithm's predictions are relatively reliable and accurate. Following Linear Regression is the AdaBoost algorithm with an $R^2$ Score of 0.35, indicating moderate explanatory power, along with a reasonable RMSE of 3.31 and MAE of 2.62. These results show that it offers reasonable predictive ability, although much less compared to Linear Regression. Logistic Regression yields an $R^2$ Score of 0.24 providing decent predictive ability. Although its RMSE of 3.63 is relatively high, it has the lowest MAE of 2.23 indicating that it minimizes prediction error relative to the target scores. Conversely, models such as Ball Tree and KDTree exhibit negative $R^2$ scores (-0.20), indicating they perform worse than a simple mean prediction. It is particularly concerning in medical applications where accurate predictions are essential. Similarly, the KNN algorithm presents a negative $R^2$ score of -0.09 and a high RMSE of 7.90, underscoring significant inaccuracies in its predictions. Models such as Random Forest and XGBoost yield low $R^2$ scores of 0.05 and 0.03 respectively indicating they struggle to capture the variability in the data, with RMSE values suggesting a considerable distance between the predicted and actual outcomes.

3. How do the features in the dataset relate to the prediction of mortality and FSS score prediction?

The initial feature set includes 103 variables spanning a wide range of medical areas, from cardiovascular records to brain computed tomography (CT) tests. Streamlining this feature set is advantageous for clinicians, as it reduces the workload and time required for collecting and managing patient data. A systematic analysis that includes investigation by a clinical researcher and iterative probing using machine learning facilitated the identification of the most significant features that contribute to the prediction of mortality and FSS scores from the original feature set. Machine learning models are known to suffer from the *curse of dimensionality* wherein an increase in features doesn't translate to an increase in the performance of the machine learning model. By systematically eliminating features that are redundant, irrelevant, and demonstrate poor performance, the team identified correlations with mortality and FSS scores, while also leveraging feature importance models derived from their analyses. The condensed feature set is developed using the feature importance models developed for mortality and FSS scores using the Matplotlib library in Python. Any feature that did not have at least a two percent relevance is eliminated from the dataset. The threshold of two percent is derived from prior work in the medical domain involving high-dimensional datasets. Furthermore, the condensed feature set is verified by a clinical researcher to ensure that all the key features are captured. Any features that are determined to be essential are added to the feature set regardless of their impact on the metrics. The rationale behind it is that although individual features may not be highly descriptive always, the combination of certain features is highly descriptive." The original feature set is condensed to 27 and 25 features for mortality and FSS score prediction respectively, as shown in Table V. Please note that the feature names shown in Table V are as observed in the original dataset. Each of the features is explained in the feature set publicly available in the Github repository (section 3.1). The condensed feature set enables clinicians to focus on the most relevant features, enhancing predictive accuracy and reducing clutter from irrelevant data. Models that operate with fewer, more pertinent features can generalize reliable predictive patterns more effectively. In the realm of medical data, this is essential, as fewer features require less computational power. Ultimately, these insights empower healthcare professionals by deepening their understanding of the clinical factors contributing to traumatic brain injury (TBI).

**Table V:** Condensed Feature Set for Mortality and FSS Score Prediction

| Mortality | FSS Score |
|---|---|
| admittoentnut | age |
| age | admittoentnut |
| cardiacarrested | cardiacarrestother |
| cardiacarrestor | cardiacarrestprehosp |
| cardiacarrestprehosp | cardiacarrestyn |
| cardiacarrestyn | ctcompress |
| ctcompress | ctepihematoma |
| ctepihematoma | ctintraparhem |
| ctintraparhem | ctintraventhem |
| ctintraventhem | ctmidlineshift |
| ctmidlineshift | ctsubhematoma |
| ctsubhematoma | decomcranyn |
| decomcranyn | entnutyn |
| entnutyn | female |
| female | gcsicu |
| gcsicu | gcsseded |
| gcsseded | hosplos |
| hosplos | icpyn1 |
| icpyn1 | injurymech |
| injurymech | newgastyn |
| lmbrdrainyn | puplrcticu |
| newgastyn | rxhypsal |
| puplrcticu | rxinotrvas |
| rxhypsal | subhemyn |
| rxinotrvas | tpnyn |
| subhemyn | |
| tpnyn | |

4. What is the clinical relevance of the results in guiding the selection of one algorithm over another?

The t-test is conducted to determine the algorithms that are best suited in a clinical setting as it is observed from RQ-1 and RQ-2 that multiple algorithms produce quality outcomes for mortality and FSS score prediction respectively [43]. The t-test helps to assess whether the performance difference between two models is meaningful or due to random chance. The test calculates a p-value, which indicates the probability that any observed difference occurred by chance. If the p-value is less than 0.05, the difference is considered statistically significant, suggesting one model performs better than the other. Conversely, if the p-value is greater than 0.05, it implies no significant difference between the models, meaning their performances are likely comparable. Six independent t-tests are performed on the top three algorithms identified for predicting mortality and FSS scores, based on their respective performance metrics. Extra Trees, Logistic Regression, and XGBoost are identified as the top three algorithms for mortality prediction. AdaBoost, Linear Regression, and Logistic Regression are identified as the top three algorithms for FSS score prediction. The results of the t-tests are shown in Table VI.

**Table VI:** T-test Results for Mortality and FSS Score Prediction

| T-test | Prediction | P-Value | Significant (yes/no) |
|---|---|---|---|
| Logistic Regression vs XGBoost | Mortality | 0.37 | No |
| Logistic Regression vs Extra Trees | Mortality | 0.82 | No |
| XGBoost vs Extra Trees | Mortality | 0.60 | No |
| Linear Regression vs AdaBoost | FSS Score | 0.0001 | Yes |
| Linear Regression vs Logistic Regression | FSS Score | 0.74 | No |
| Logistic Regression vs AdaBoost | FSS Score | 0.89 | No |

For mortality prediction, the high p-values suggest that none of the algorithms exhibit a statistically significant advantage over the others. The observed differences between the algorithms are likely due to random variability in the data. Like mortality prediction, the p-values for FSS score prediction are high except for the p-value computed by the Linear Regression vs AdaBoost t-test. This indicates that there is a significant difference between the performance of the two models revealing that AdaBoost possesses distinct predictive power from Linear Regression, but not from Logistic Regression. Furthermore, it suggests that Logistic Regression and Linear Regression are comparable in performance. A non-significant p-value doesn't mean the models are identical, but rather that the observed performance differences are not large enough to rule out chance. It suggests that all compared algorithms are equally effective based on the current dataset and experimental conditions. Considering the algorithms are equally effective, the following factors are to be considered as an algorithm is chosen in a clinical setting:

a) **Model Complexity:** Model complexity refers to how intricate a model's structure is. Simpler models, like linear regression and logistic regression, rely on linear relationships, while more complex models, such as AdaBoost, XGBoost or Extra Trees, capture non-linear patterns through ensembles of decision trees. Complex models can often achieve better performance but require more data and computational resources.

b). **Interpretability:** Interpretability is the ease with which humans can understand how a model makes predictions. Algorithms such as Linear Regression and Logistic Regression are highly interpretable since their coefficients show the direct impact of each feature. In contrast, XGBoost and Extra Trees are harder to interpret due to their complex decision paths and interactions across many trees.

c). **Parameter Tuning:** Parameter tuning involves finding the optimal set of hyperparameters that improve a model's performance. Simple models like Linear Regression and Logistic regression require fewer hyperparameters to tune, while complex models, such as AdaBoost and XGBoost, demand extensive tuning (e.g., learning rate, tree depth) to achieve optimal performance.

d). **Training Time:** Training time is the amount of time a model takes to learn from the data. Simpler models, like logistic regression, train quickly, especially on smaller datasets. However, more complex models like XGBoost or Extra Trees take longer due to the iterative processes involved in building and optimizing multiple trees.

## IV. Future Work and Conclusion

TBI is a complex condition whose diagnosis relies on multiple factors. The current work demonstrates the feasibility of machine learning algorithms for the prediction of mortality and FSS scores that empower clinicians to predict the outcome of TBI patients. In addition to employing machine learning techniques, collaborating with clinical researchers to interpret model predictions and refine algorithms greatly helps in the development of precise machine learning models for TBI. It ensures that models are accurate and align with clinical needs. Integrating these models into clinical workflows provides valuable decision support, potentially improving patient outcomes by enabling more personalized and timely interventions.

Future research will aim to further develop these models using advanced machine learning methods, such as deep learning and ensemble techniques, to enhance their predictive performance and clinical utility. Incorporating these methods is expected to improve the models' ability to handle complex, high-dimensional data and capture subtle patterns that may be missed by traditional approaches. Expanding the dataset to include patient care data from a larger corpus of patients will ensure robustness and mitigate the influence of randomness and biases potentially contained in the current dataset. This will involve integrating data from multiple institutions and covering a broader range of patient demographics and clinical scenarios. Reducing noise within the model implementations, through techniques such as feature selection and data augmentation, will ultimately help increase the models' predictive accuracy.

## Acknowledgment

The authors express their gratitude to the University of Colorado School of Medicine for providing the dataset on TBI as part of the Harmonized Pediatric Traumatic Brain Injury Hackathon via their Github repository.